\pdfoutput=1

\documentclass[11pt]{article}

\usepackage{EMNLP2023}

\usepackage{times}
\usepackage{latexsym}
\usepackage{graphicx}
\usepackage[T1]{fontenc}

\usepackage[utf8]{inputenc}
\usepackage{booktabs}
\usepackage{microtype}

\usepackage{inconsolata}

\title{GenQuest: An LLM-based Text Adventure Game for Language Learners}

\author{
Qiao Wang \\
Hosei University, Japan \\
\texttt{judy.wang@hosei.ac.jp}
\And
Adnan Labib \\
King's College London, UK \\
\texttt{adnan.1.labib@kcl.ac.uk}
\AND   
Robert Swier \\
Kindai University, Japan \\
\texttt{robert.swier@lac.kindai.ac.jp}
\And
Michael Hofmeyr \\
Tokyo Uni. of Science, Japan \\
\texttt{hofmeyr@rs.tus.ac.jp}
\And
Zheng Yuan \\
University of Sheffield, UK \\
\texttt{zheng.yuan1@sheffield.ac.uk}
}

\begin{document}
\maketitle
\begin{abstract}
GenQuest is a generative text adventure game that leverages Large Language Models (LLMs) to facilitate second language learning through immersive, interactive storytelling. The system engages English as a Foreign Language (EFL) learners in a collaborative “choose-your-own-adventure” style narrative, dynamically generated in response to learner choices. Game mechanics such as branching decision points and story milestones are incorporated to maintain narrative coherence while allowing learner-driven plot development. Key pedagogical features include content generation tailored to each learner’s proficiency level, and a vocabulary assistant that provides in-context explanations of learner-queried text strings, ranging from words and phrases to sentences. Findings from a pilot study with university EFL students in China indicate promising vocabulary gains and positive user perceptions. Also discussed are suggestions from participants regarding the narrative length and quality, and the request for multi-modal content such as illustrations.
\end{abstract}

\section{Introduction}
Digital games have proven effective in enhancing the motivation and learning outcomes of second language (L2) learners. Among the numerous genres of games, narrative-based adventure games, ranging from the purely text-based those incorporating 3D graphics and virtual realty, can harness storytelling to engage learners with authentic L2 interaction. In these games, players take an active role in a branching story, reading and making choices in the target language. This can increase learner motivation and contextualize reading practice, leading to improved reading engagement and incidental vocabulary acquisition in L2 environments \citep{Reinders2015, Dixon2022, Chowdhury2024}. However, traditional narrative games rely on pre-authored, static storylines featuring a finite set of branches, which cannot adapt to individual interests or proficiency levels. Such a lack of personalization may limit their pedagogical effectiveness \citep{Zelinka2018, Risi2020}. 

Recent advances in Large Language Models (LLMs) offer promising new avenues to explore in the design of text-based narrative games. Modern LLMs (e.g., GPT-style models) can generate coherent story continuations in response to player input, enabling virtually unlimited branching narratives and player-driven plot development \citep{Harshitha2022, Lanzi2023}. Additionally, LLMs can tailor linguistic complexity to a player's skill level. For example, LLMs can rephrase descriptions or modify vocabulary difficulty to create a personalized and level-appropriate reading experience \citep{Harshitha2022, Jegede2024}. To leverage this powerful new potential, we have created  \textit{GenQuest}, a novel text-based narrative game that generates a branching narrative adventure compatible with each player's language proficiency while providing in-story language support. 


\begin{figure*}[t]
    \centering
    \includegraphics[width=1\linewidth]{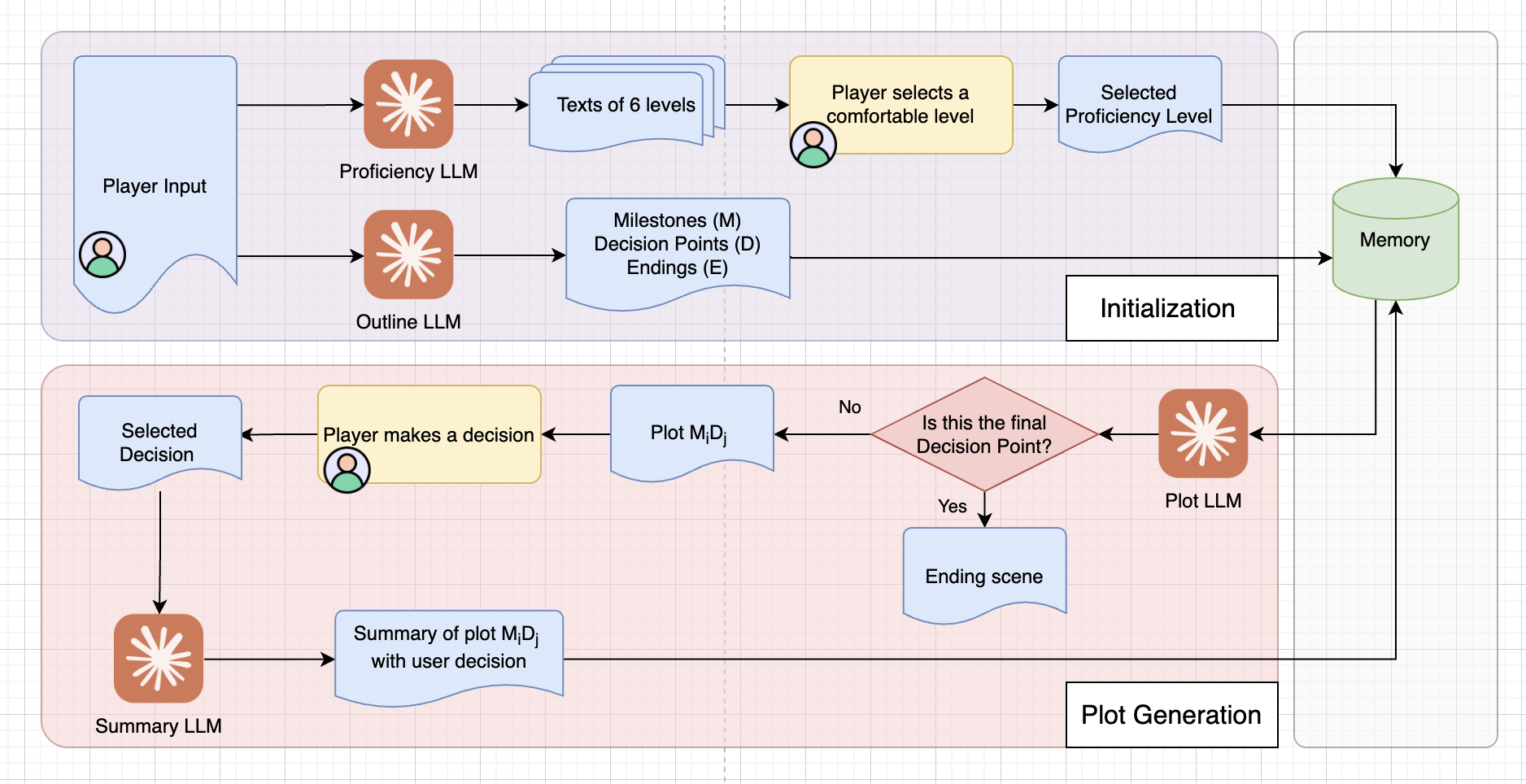}
    \caption{Architecture and workflow of the story module}
    \label{fig:story module}
\end{figure*}

\section{Background}

Text adventure games, also called interactive fiction (IF), have their roots in branching narratives such as the \textit{Choose Your Own Adventure} books originally published by Bantam Books from the late-1970s, where readers shaped the path of the narrative by selecting from pre-written options. With the rise of personal computers, text adventure games like \textit{Zork} \citep{zork} introduced parser-based input, enabling players to type commands and explore worlds with rich story lines. Such early works of IF established conventions for world exploration, puzzles, and quest structures that remain influential today. However, both paper-based and early digital IF works were constrained by fixed storylines and heavy authoring demands.

To address these constraints, researchers have developed computational systems for automated narrative generation. Work on quest generation \citep{ammanabrolu2020a} and world-building from knowledge graphs \citep{ammanabrolu2020b} aimed to expand re-playability and coherence. Environments such as \textit{TextWorld} \citep{cote2018} and \textit{ScienceWorld} \citep{wang2022} provided controllable testbeds for training and evaluating agents in text-driven settings. Role-playing datasets such as \textit{LIGHT} \citep{urbanek2019,fan2019} and the large corpus of \textit{Dungeons \& Dragons} gameplay \citep{callisonburch2022,zhu2023} reframed IF as dialogue with state-tracking problems. With the advent of LLMs, the field has shifted from handcrafted pipelines to generative storytelling. LLMs have been shown to encode implicit representations of entities and meaning \citep{li2021} in order to sustain coherent dialogue and action in complex environments \citep{park2023,bubeck2023}. 

Parallel to these technical advances, interactive environments have been increasingly applied in educational research. Studies on inquiry-based tasks \citep{yuan2019,tamari2019} and scientific reasoning in games \citep{wang2022} demonstrates how narrative- or task-oriented interactions can serve as effective mechanisms for fostering comprehension and problem-solving. Grounded cognition theory \citep{barsalou2008} and embodied approaches to meaning \citep{feldmannarayanan2004} support the idea that learning is enhanced when language is situated in meaningful, action-driven contexts. Improvisational storytelling with an AI collaborator further demonstrates how narrative co-creation can foster engagement and creativity in educational settings \citep{martin2016,martin2017,mathewson2017,mirowski2019}.

Despite their alignment with contextualized learning, relatively few systems have directly applied text adventure games to second language acquisition (SLA) \citep{wright2024}. Narrative gameplay naturally provides repeated exposure to input, in-context vocabulary, and built-in comprehension checks through meaningful decision-making, yet most applications have focused on gamification more broadly rather than branching narratives specifically (e.g., \citet{thurairasu2022}). Existing work identifies several challenges particularly relevant to L2 learners: maintaining narrative coherence across branches, providing effective language scaffolding, and calibrating linguistic difficulty to individual proficiency levels. The shift toward LLM-driven narratives offers opportunities to address these issues: quests and world models can support coherence, dynamic glossing can scaffold vocabulary learning, and adaptive generation can align text with learners’ needs. In spite of this strong educational potential, the application of text adventure games specifically for language learning remains underexplored.

\section{Methodology}
\label{methods}
To leverage the potential of LLM-powered branching narratives in language learning, we developed \textit{GenQuest}, a web-based text adventure game that makes language learning more enjoyable and engaging. The system architecture integrates two commercial LLMs, GPT-4o\footnote{\url{https://openai.com/index/hello-gpt-4o/}} and Claude Sonnet 3.7\footnote{\url{https://www.anthropic.com/news/claude-3-7-sonnet}}. 
During development, several recent LLMs were tested, including DeepSeek V3 \footnote{\url{https://api-docs.deepseek.com/news/news250325}}, LLaMA-3 \footnote{\url{https://ai.meta.com/blog/meta-llama-3/}}, GPT-4o, and Claude 3.7. Comparative evaluation suggested that Claude 3.7 was more effective for story generation, as it produced narratives with more engaging plot twists and provided better text alignment with targeted language levels. By contrast, GPT-4o proved stronger for generating clear, context-sensitive language explanations. The role of two LLMs in different parts of system will be described in later in this section. The backend is built in Python with a RESTful API \footnote{\url{https://github.com/microsoft/api-guidelines}} and communicates with a Vue.js based front end \footnote{\url{https://v2.vuejs.org/}}. The source code is available at \url{https://github.com/judywq/ai-text-game}. A web-based user interface is hosted at \url{https://atg.judywang.jp/}.

\textit{GenQuest} integrates both game-oriented and pedagogical features into a unified framework. On the game side, the system provides a wide variety of predetermined narrative genres and interactive gameplay through player input (at initialization) and player choice (at decision points). On the pedagogical side, the system aligns all content to the learner’s language proficiency level, allows the player to query unknown text strings in context, and stores all queried items for later review. These features are realized through two interconnected modules: a story module and a language module.

\subsection{Story module}
The story module consists of two pipelines that share a common memory system: an initialization pipeline and a plot generation pipeline. Figure~\ref{fig:story module} illustrates the architecture and workflow.

\begin{figure}[t]
    \centering
    \includegraphics[width=1\linewidth]{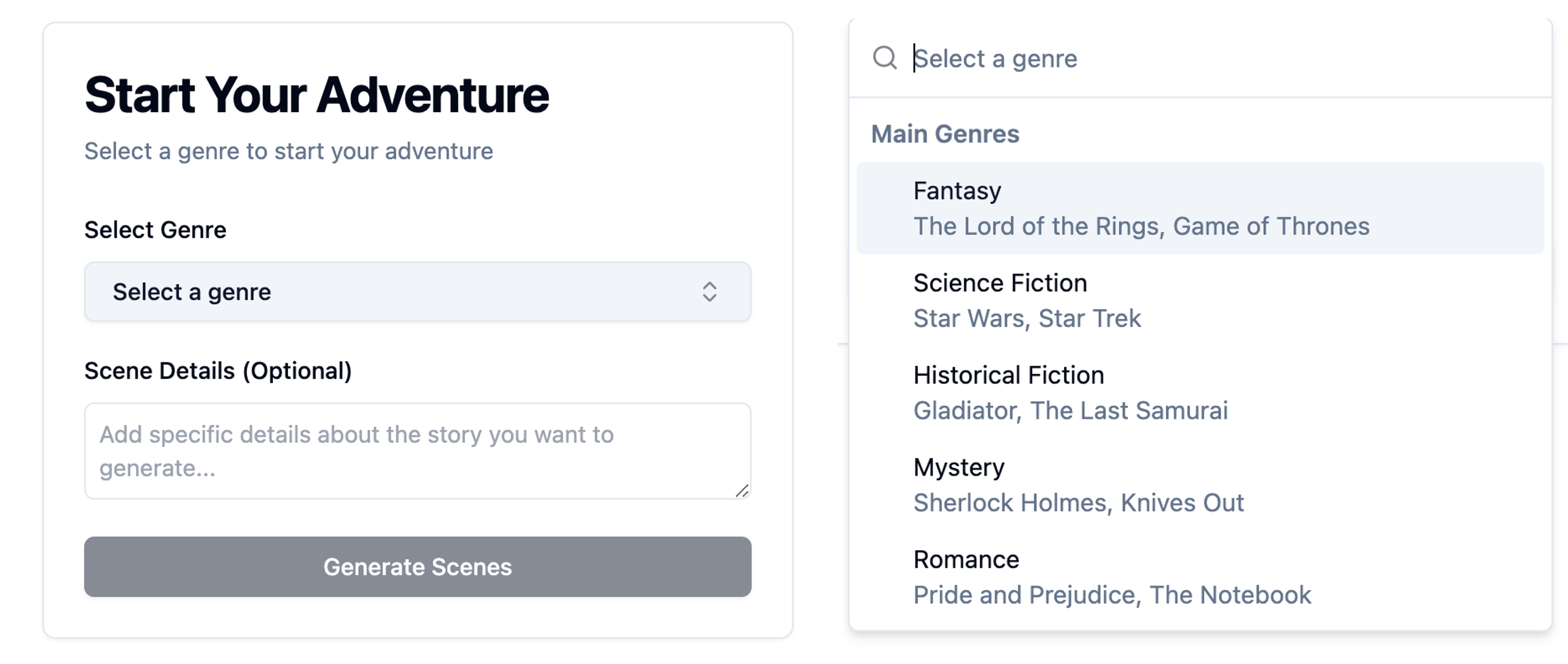}
    \caption{User input at game initiation}
    \label{fig:user input}
\end{figure}

\paragraph{Initialization pipeline.}
The initialization pipeline begins with user input (see Figure \ref{fig:user input}). The player first selects a narrative genre from a set of pre-defined options (e.g., fantasy, mystery, science fiction), which are accompanied by exemplative cinematic works to illustrate the choice. Players may provide additional input (e.g., specifying characters or settings) or leave the content creation entirely to the system. 

\begin{figure*}[t]
    \centering
    \includegraphics[width=1\linewidth]{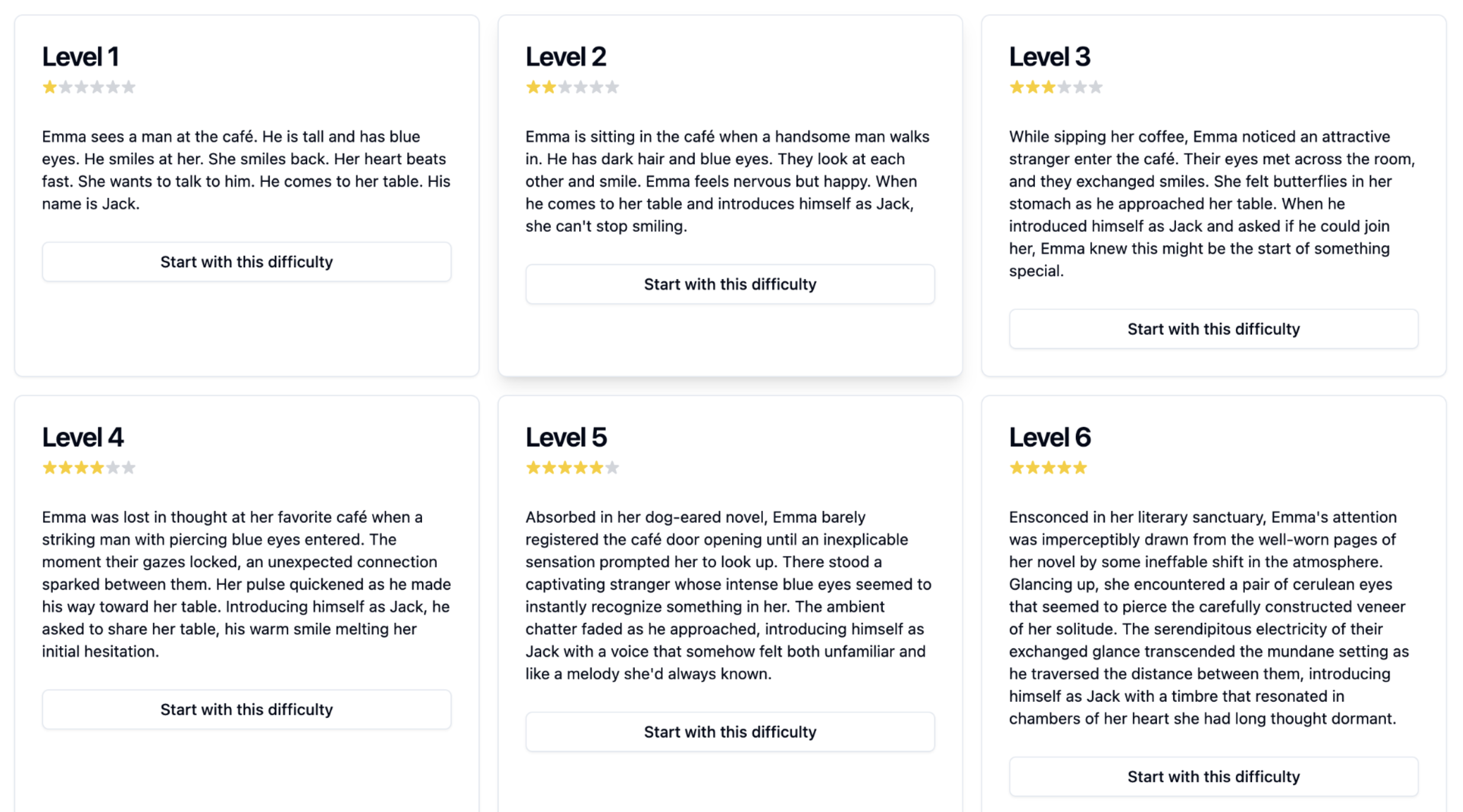}
    \caption{Example texts for six CEFR levels.}
    \label{fig:CEFR levels}
\end{figure*}

Next, the proficiency LLM, based on Claude 3.7, generates six semantically similar narrative samples ranging from level A1 to C2 of the Common European Framework of References for Languages (CEFR) (see Figure \ref{fig:CEFR levels} for an example). The player selects the version they find most appropriate. Concurrently, the outline LLM, based on Claude 3.7, generates a structured narrative outline consisting of milestones (M), decision points (D), and possible endings (E). The number of milestones, decision points, and endings can be configured in the backend. Both the selected proficiency level and the narrative outline are stored in the system memory for use in subsequent stages.

\paragraph{Milestones and decision points.}
To balance narrative freedom and coherence, the outline is structured around a sequence of milestones, or checkpoints representing key story events, that the player encounters regardless of their previous choices. Between milestones, the story branches into multiple decision points, where players select from a list of actions that affect how the story progresses toward the next milestone. This structure was inspired by narrative design in commercial interactive titles such as \textit{Detroit: Become Human}\footnote{\url{https://www.quanticdream.com/en/detroit-become-human}}, which uses converging checkpoints to prevent exponential growth of branching paths while preserving player agency. In \textit{GenQuest}, this approach ensures that learners’ decisions meaningfully shape the narrative, yet all players experience core story events. 

\paragraph{Plot generation pipeline.}
Once the story begins, the plot LLM, based on Claude 3.7, retrieves information from the memory and generates the first story segment (plot $M_iD_j$), conditioned on the structured outline and the player’s proficiency level. The system then presents a decision point, at which point the player selects an action. After each decision, the summary LLM, based on Claude 3.7, condenses the current plot segment and appends it to the memory system, ensuring continuity across branches. This loop---story generation, decision-making, and summarization---continues until the player reaches the final decision point of the last milestone. At this stage, the plot LLM produces an ending consistent with the player’s cumulative choices.

\subsection{Language module}
\begin{figure}[t]
    \centering
    \includegraphics[width=1\linewidth]{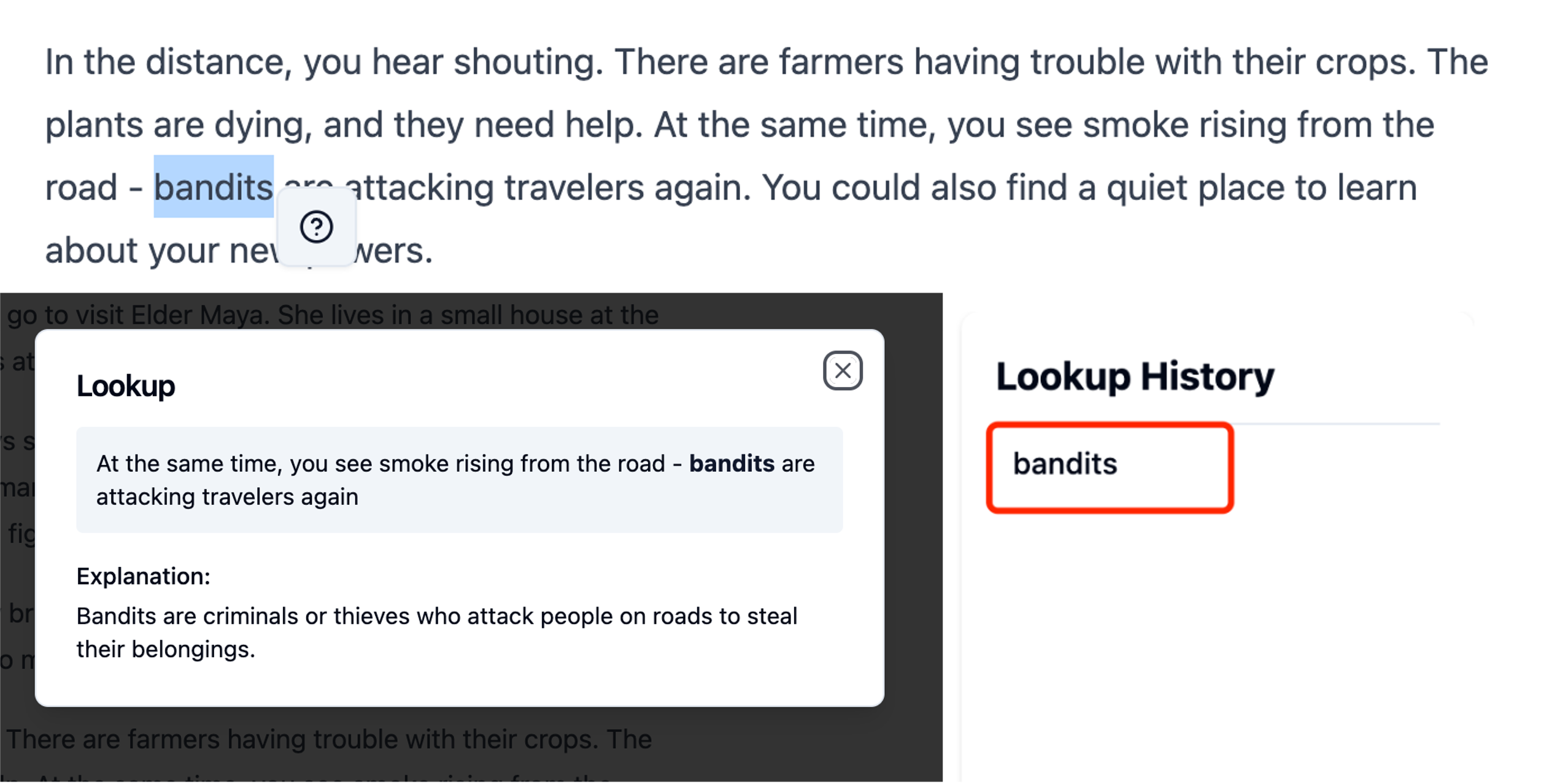}
    \caption{Language query and query history}
    \label{fig:query}
\end{figure}

The language module provides in-game scaffolding for language learning (see Figure \ref{fig:query}). The user interface (UI) allows a player to highlight any text string---i.e., words, phrases and even sentences---during gameplay, after which a query icon in the form of question mark will appear. Clicking the icon triggers the language assistant, which is based on GPT-4o. The assistant then generates a contextually appropriate explanation of the string at the player’s proficiency level. Each queried string is also simultaneously stored in a personal list for later review.

\section{Pilot Study}
This pedagogical features of \textit{GenQuest} can facilitate reading comprehension and vocabulary learning through context. As such, we conducted a pilot study in the form of a user experiment to evaluate the effectiveness of the system on learners' vocabulary learning. The research questions are as follows:

RQ1: Can the game help learners acquire new vocabulary in context?

RQ2: What are learner perceptions of the game?

\subsection{Participants and procedures}
 Ten second-year undergraduate students at a university in China were recruited for the pilot study. One student withdrew during the study, resulting in a total of nine participants. Based on their scores from the College English Test Level 4, the participants ranged from A2 to B1 in CEFR-benchmarked proficiency levels. All participants provided informed consent and received instructions on how to use the system. The study lasted for five consecutive days. Each participant was asked to complete one session per day, resulting in five sessions per participant. At the end of the five-day period, participants completed a vocabulary test and a short survey. 

\subsection{Instruments}

\paragraph{Vocabulary test.}
A vocabulary test was used to answer RQ1 and was tailored to each participant. We downloaded from the system logs all the text strings that each participant looked up and selected 20 words of the highest relevance and frequency for each gameplay session. The tests were administered online. For each of the 20 words, participants were first asked to indicate whether they knew its meaning. If they selected “yes,” they were required to type the meaning either in English or Chinese. Responses were scored by two human raters according to the following scheme: 1 point for a correct meaning, 0.5 points for a partially correct or approximate meaning, and 0 points for an incorrect meaning or a “no.” Inter-rater disagreements were discussed until consensus was reached. The maximum score per participant was 20.  

\paragraph{Survey.}
After the vocabulary test, participants completed a survey based on the Technology Acceptance Model \citep[TAM]{davis1989user}. The questionnaire contained 12 Likert-scale items, with six items targeting Perceived Usefulness (PU) and six targeting Perceived Ease of Use (PEOU). Each item was rated on a seven-point Likert scale (1 = strongly disagree, 7 = strongly agree). In addition, an open-ended question asked participants to provide suggestions or general comments about the game. The Liker-scale questions were analyzed quantitatively while the thematic coding was performed on the open-response question. The full survey is included in Appendix~\ref{appendix:survey}.

\subsection{Results}

\subsubsection{RQ1: Vocabulary test results}
Table~\ref{tab:vocab-results} shows the vocabulary test scores of the nine participants.

\begin{table}[t]
\centering
\begin{tabular}{l c}
\hline
\textbf{Participant} & \textbf{Score (/20)} \\
\hline
1 & 17.5 \\
2 & 9 \\
3 & 13 \\
4 & 13 \\
5 & 9 \\
6 & 17 \\
7 & 18 \\
8 & 6 \\
9 & 18.5 \\
\hline
\textbf{Average (M)} & 13.44 \\
\textbf{SD} & 4.62 \\
\hline
\end{tabular}
\caption{Vocabulary test results of the nine participants.}
\label{tab:vocab-results}
\end{table}

\begin{table*}[t]
\centering
\begin{tabular}{llp{9.4cm}cc}
\toprule
\textbf{Construct} & \textbf{Item} & \textbf{Statement} & \textbf{Mean} & \textbf{SD} \\
\midrule
PU & PU1 & Using the text-adventure game helps me learn new vocabulary more quickly. & 5.00 & 1.41 \\
PU & PU2 & The game improves my performance in vocabulary learning. & 5.22 & 1.09 \\
PU & PU3 & Using the game increases my vocabulary. & 5.56 & 1.13 \\
PU & PU4 & The game enhances my effectiveness in learning new words. & 5.22 & 1.39 \\
PU & PU5 & The game makes learning vocabulary easier for me. & 5.44 & 1.24 \\
PU & PU6 & I find the game useful for improving my English vocabulary. & 5.56 & 0.88 \\
\midrule
PEOU & PEOU1 & Learning how to play the game was easy for me. & 5.44 & 1.24 \\
PEOU & PEOU2 & I find it easy to use the game to learn vocabulary. & 5.33 & 1.50 \\
PEOU & PEOU3 & My interaction with the game is clear and understandable. & 6.00 & 1.12 \\
PEOU & PEOU4 & The game interface is easy to understand. & 6.00 & 0.87 \\
PEOU & PEOU5 & It is easy for me to become skillful at playing the game. & 5.89 & 0.93 \\
PEOU & PEOU6 & I find the game easy to use overall. & 6.44 & 0.53 \\
\midrule
\multicolumn{3}{r}{\textbf{Construct summary (per-participant average across items)}} & \textbf{Mean} & \textbf{SD} \\
\multicolumn{3}{r}{Perceived Usefulness (PU; items 1--6)} & 5.33 & 1.02 \\
\multicolumn{3}{r}{Perceived Ease of Use (PEOU; items 7--12)} & 5.85 & 0.81 \\
\bottomrule
\end{tabular}
\caption{Descriptive statistics for 12 Likert items}
\label{tab:tam-12items}
\end{table*}

Participants' scores ranged from 6 to 18.5 out of 20, with an average of 13.44. This suggests that on average, participants learned or retained around two-thirds of the vocabulary items they had looked up during gameplay. Despite the small sample size, there is encouraging evidence that interactive narrative gameplay supported vocabulary acquisition and retention. A closer look at participants who had the lowest scores: Participant two (played four stories at A2 and one at B1), participant five (four at A2 and one at B1), and participant eight (three at A2 and two at B1), and participants who had the highest scores: Participants one (four at B2 and one at B1), participant seven (three at A2 and two at B1), and participant nine (three at A2 and two at A1), revealed that the lowest scores were generally associated with lower chosen proficiency levels. These initial results suggest that participants who selected easier texts were themselves likely at lower proficiency, and their baseline language level influenced their learning outcomes. However, there was one notable exception: Participant nine chose lower-level texts (A1–A2) but still achieved one of the highest scores. This points to the complex role of both learner proficiency and text difficulty in shaping vocabulary learning outcomes.

\subsubsection{RQ2: Survey results}
\paragraph{Quantitative results.}

Table~\ref{tab:tam-12items} presents the descriptive statistics of the 12 questions. To examine internal reliability, Cronbach’s $\alpha$ was calculated for each construct. The PU items demonstrated excellent internal consistency ($\alpha$ = 0.92), while the PEOU items showed good reliability ($\alpha$ = 0.84). 

Overall, participants reported moderately positive views of the system’s usefulness. The PU scale yielded an average rating of 5.33 (SD = 1.02), with the highest-rated item being “I find the game useful for improving my English vocabulary” (mean = 5.56). Meanwhile, the PEOU scale averaged 5.85 (SD = 0.81), with particularly strong agreement on items such as “I find the game easy to use overall” (mean = 6.44) and “The game interface is easy to understand” (mean = 6.00). These findings suggest that most learners considered the game useful for vocabulary learning and easy to use.

\paragraph{Qualitative results.}

Analysis of the open-ended responses revealed several recurring themes regarding learners’ perceptions of the game system. Overall, participants acknowledged the novelty and motivational value of interactive storytelling, but also identified specific areas for improvement. Table \ref{tab:qualitative-counts} presents the themes with instance counts based on participant comments.
\begin{table*}[t]
\centering
\begin{tabular}{l c}
\hline
\textbf{Theme} & \textbf{Number of Participants} \\
& \textbf{(N=9)} \\
\hline
Usefulness and engagement & 3 \\
Vocabulary support & 5 \\
Difficulty adjustment & 2 \\
Narrative quality and coherence & 3 \\
User experience and interactivity & 3 \\
Suggestions for enhancement (e.g., recycling vocab, fanfiction prompts) & 2 \\
\hline
\end{tabular}
\caption{Themes emerging from open-ended survey responses.}
\label{tab:qualitative-counts}
\end{table*}

Under Usefulness and Engagement,
learners emphasized the game’s novelty and its potential for motivating vocabulary learning. For instance, one participant described it as ``very novel, suitable for learners at different proficiency levels, and helpful for improving English reading ability in a short time''. Others noted that the variety of story genres and multiple endings made the reading experience enjoyable and sustained their interest.

Under Vocabulary Support,
the in-game language query function was considered valuable, but participants also highlighted its limitations. One participant reported that English-only explanations sometimes contained other unfamiliar words, making comprehension difficult, and suggested adding first language (L1) translations---Chinese in this case---as an option. Another proposed expanding explanations to include collocations, multiple senses, or usage examples, noting that such enhancements would improve learning efficiency. Some also recommended recycling queried words in later stories to reinforce retention.

Under Difficulty Adjustment,
participants expressed a desire for finer-grained control of text difficulty. While the proficiency level selection in user input was appreciated, one participant suggested that vocabulary knowledge should be explicitly assessed first and used to assign more suitable levels. Another participant emphasized the need to subdivide proficiency levels, particularly for the B1–B2 range, as they fell within this range but still encountered uneven difficulty depending on story genre.

Under Narrative Quality and Coherence,
although the branching stories were generally engaging, several participants observed issues with narrative coherence. Examples included logical inconsistencies between story branches and overly lengthy passages that reduced readability. Some felt that the AI-generated stories lacked background detail or relied on clichéd tropes. Interestingly, one participant found greater engagement when providing the AI with familiar story frameworks (e.g., fanfiction scenarios) in user input, which improved character consistency and reader investment.

Under User Experience and Interactivity,
one participant noted the inconvenience of being forced to wait for more than one minute for a story to initiate. Others emphasized that ``an all-text interface can easily discourage users'' and recommended integrating illustrations to improve accessibility. More interactive mechanics were also suggested, such as fill-in-the-blank options where learners could actively supply vocabulary to direct character actions, or direct text input as opposed to pre-determined action lists.

\section{Discussions}

The pilot study shows that LLM-powered text adventures hold promise for enhancing vocabulary learning, while also raising important design questions around difficulty calibration, multimodal engagement, and vocabulary scaffolding. Addressing these issues will be critical for building more robust, scalable, and pedagogically effective systems.

First, the alignment of CEFR levels remains challenging. While the game can generate texts from A1 to C2, higher-level texts often contain rare or specialized vocabulary that may not be immediately useful for learners. In other words, LLMs may not be able to accurately generate texts of a certain CEFR level, which has been mentioned by previous studies as well \cite{wang2024evaluation}. More careful curation of lexical content, or the integration of frequency-based filters, may help ensure that advanced texts remain pedagogically relevant.  

Second, regarding the need for multimodal enhancements such as images or simple visualizations, integrating image-generation models would be a promising direction to explore in the future. However, ensuring consistency across generated visuals and latency in generation remain a significant technological constraint \citep{hollein2024three,google2024mobilediffusion}.  

Third, some learners expressed difficulty with English-only explanations and requested the use of their L1 for clarification. This raises the question of how much information should be provided in real time. Selectively including bilingual glosses, or integrating a dictionary API for deeper exploration outside of gameplay, could balance immediate support with opportunities for later consolidation.

Apart from the above improvements, in future work, a quiz system can also be added, including vocabulary or reading comprehension questions, to expand the system beyond learning to testing.

\section{Conclusions}

This study explored the design and evaluation of\textit{ GenQuest}, an LLM–powered text adventure game for language learning. The results of a five-day pilot suggest that interactive narrative gameplay can effectively support vocabulary acquisition and retention. Survey responses further indicated positive learner perceptions regarding the usefulness and ease of use of the game. At the same time, the study revealed several aspect in which it can still be improved. Learners suggested that vocabulary explanations should be simplified or supplemented with translations, that narrative coherence occasionally faltered, and that usability would benefit from multi-modal enhancements such as illustrations. Although the game currently supports only English, the framework of the game can be adapted to other languages using backbone LLMs that support relevant target language(s).

\section*{Limitations}

This study has several limitations that should be acknowledged. First, the sample size was small (nine participants), which restricts the generalizability of the findings. Second, the experiment was conducted over a short period of five days, limiting the ability to assess long-term vocabulary retention and sustained engagement. Third, all participants were Chinese university students, and cultural and educational factors specific to this group may limit the applicability of results to other learner populations. Additionally, the evaluation focused primarily on vocabulary outcomes and self-reported perceptions, while other aspects such as reading fluency, narrative comprehension, and long-term motivational effects were not systematically assessed. Finally, the system itself is still a prototype, and technical limitations in narrative coherence, scaffolding, and interface design may have influenced participants’ experiences. Future research with larger and more diverse samples, extended study durations, and broader evaluation metrics will be necessary to validate and extend these findings.

\section*{Ethics Statement}

The study was conducted in accordance with institutional research ethics guidelines. Informed consent was obtained from all participants, and they were explicitly told that participation was voluntary and unrelated to their course grade or academic evaluation. Participants were free to withdraw from the study at any time without penalty. All personal data, including gameplay logs and vocabulary test results, were anonymized before analysis to protect privacy. No sensitive personal information was collected, and care was taken to ensure that the game content avoided harmful, biased, or culturally inappropriate material. The  source code and game UI are made available solely for research purposes, and double-blind procedures were followed during review.  

\section*{Acknowledgements}
We would like to thank the 9 participants for their participation in the pilot study and their teacher, Dr. Ke LI, for undertaking liaison work between the researchers and the participants.

\bibliography{anthology,custom}
\bibliographystyle{acl_natbib}



\section*{Appendices}
\appendix






\section{Questionnaire}

\label{appendix:survey}

I. Likert-scale questions:

For each of the following statement, please rate from 1 to 7 to show how much you agree or disagree. 

1 = Strongly disagree, 2 = Disagree, 3 = Somewhat disagree, 4 = Neutral (neither agree nor disagree), 5 = Somewhat agree, 6 = Agree, 7 = Strongly agree.
\begin{enumerate}
    \item Using the text-adventure game helps me learn new vocabulary more quickly.
    \item The game improves my performance in vocabulary learning.
    \item Using the game increases my vocabulary.
    \item The game enhances my effectiveness in learning new words.
    \item The game makes learning vocabulary easier for me.
    \item I find the game useful for improving my English vocabulary.
    \item Learning how to play the game was easy for me.
    \item I find it easy to use the game to learn vocabulary.
    \item My interaction with the game is clear and understandable.
    \item The game interface is easy to understand.
    \item It is easy for me to become skillful at playing the game.
    \item I find the game easy to use overall.

\end{enumerate}
II. Free-response question:

Please provide any suggestions or opinions you have about the game.
\end{document}